%% file: nbiig.tex
\documentclass[letterpaper]{article} 
\usepackage{aaai23}  
\usepackage{times}  
\usepackage{helvet}  
\usepackage{courier}  
\usepackage[hyphens]{url}  
\usepackage{graphicx} 
\usepackage{natbib}  
\usepackage{caption} 
\usepackage{algorithm}
\usepackage{algorithmic}

\usepackage{newfloat}
\usepackage{listings}
\DeclareCaptionStyle{ruled}{labelfont=normalfont,labelsep=colon,strut=off} 
\floatstyle{ruled}
\newfloat{listing}{tb}{lst}{}
\floatname{listing}{Listing}
\title{n\textsc{BIIG}: A Neural BI Insights Generation System for Table Reporting}
\author{Yotam Perlitz, 
Dafna Sheinwald, Noam Slonim, Michal Shmueli-Scheuer\\
  IBM Research\\
  yotam.perlitz@ibm.com, \{dafna, noams, shmueli\}@il.ibm.com}

\usepackage{xcolor}
\usepackage{soul}
\usepackage{float}

\begin{document}

\maketitle

\begin{abstract}
We present n\textsc{BIIG}, a neural Business Intelligence (BI) Insights Generation system. Given a table, our system applies various analyses to create corresponding RDF representations, and then uses a neural model to generate fluent textual insights out of these representations. The generated insights can be used by an analyst, via a human-in-the-loop paradigm, to enhance the task of creating compelling table reports. The underlying generative neural model is trained over large and carefully distilled data, curated from multiple BI domains. Thus, the system can generate faithful and fluent insights over open-domain tables, making it practical and useful. 

\end{abstract}

\input 1_introduction
\input 3_system
\input 4_discussion
\input 7_acknowledgments

\bibliography{nbiig}

\end{document}

%% file: 1_introduction.tex
\section{Introduction}
\label{introduction}

\begin{figure}[t]
\begin{center}
  \includegraphics[width=1\columnwidth]{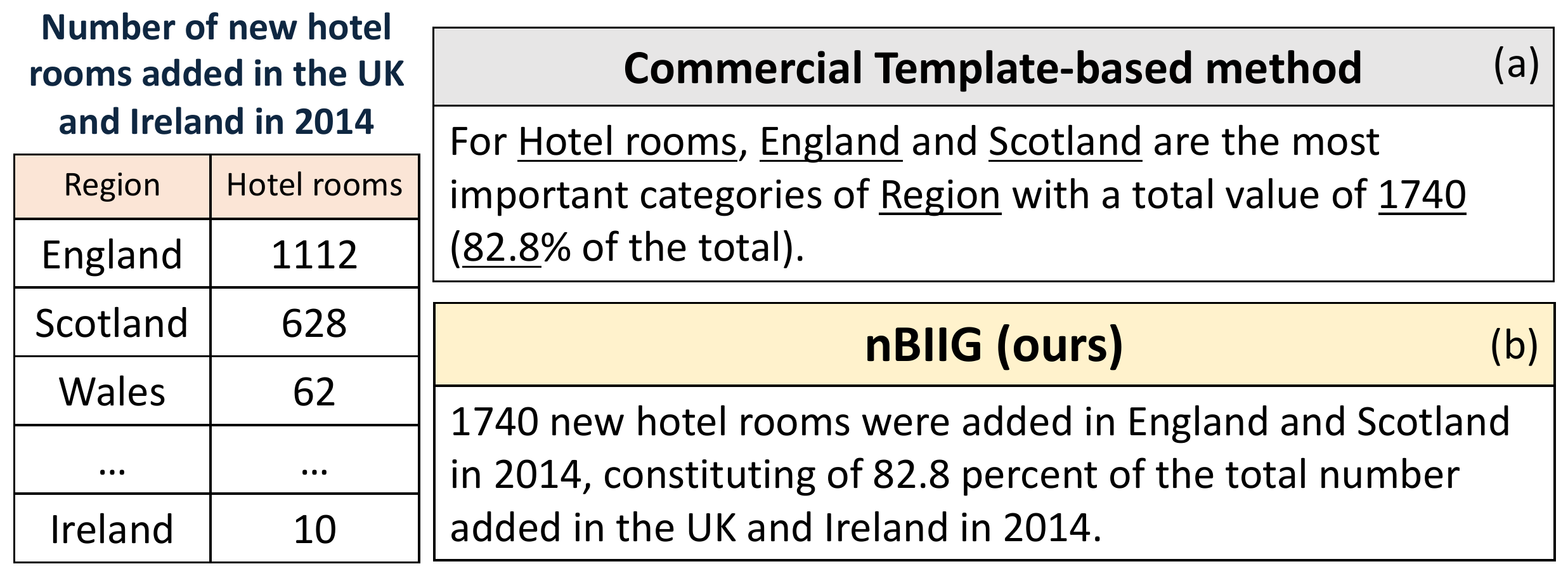}
 \end{center}
  \caption{\textbf{n\textsc{BIIG} Vs. Template-based insight generation:} Insights were generated automatically using (a) a commercial template-based method and (b) n\textsc{BIIG}. Both insights convey the same information, however, n\textsc{BIIG}s higher level of fluency makes the insight more readable and engaging.}~\label{fig:process-example}
\end{figure}

BI tools are commonly used in various industries 
with 48\% of all organizations consider them either ``critical'' or ``very important'' to their operations\footnotemark[1].
\textit{Table reports}, a natural language summary of 
the results reported in a table, 
is a prominent part in BI solutions, 
and generating 
these constitutes 
an important part of the analyst work. 
Automatic \textit{insight generation} is often used to enhance this task, by applying various analyses over the table and expressing their outcome in natural language sentences, that represent candidate insights for the analyst, to include in the report. 
For these insights to be useful, they must satisfy: 
(1) \textit{faithfulness} -- be consistent with the table; 
and (2) \textit{fluency} -- be  textually fluent. 
Current commercial products adopt a two-stage method for insight generation \cite{wang2019datashot}, implemented by two modules: 
(1) \textit{Analytics}: applying a set of statistical analyses to the data, resulting in meaning representation, and (2) Template-based \textit{surface realization}: using templates and rules to transform the analyses from their meaning representation 
into natural language insights. 
Due 
to the deterministic character of the analytics module, these methods excel in their high 
level 
of faithfulness, however, their use of template-based surface realization undermines their fluency (see Figure~\ref{fig:process-example}) which impairs their usability, as the analyst is required to re-write before any subsequent use.
Recently, End-to-End approaches for insights generation have evolved to train a generative language-model to produce insights directly from the table. 
These methods have shown to generate fluent and diverse texts \cite{perlitzDiversity2022}, but, since these generative models often struggle with numerical operations \cite{ravichander2019equate}, they 
result with 
low levels of faithfulness~\cite{chen2020logic2text}, where every second insight is unfaithful to the underlying data, deeming them unsuitable for commercial applications. 
Here, we present a neural BI Insight Generation (n\textsc{BIIG}) system, that 
maintains the above two-stage approach, 
but replaces the templates and rules of the second stage with a neural model. As shown in our evaluations, this 
enables the system to satisfy both high levels of fluency and faithfulness.
The generative model is trained on a new large-scale data from multiple BI domains to ensure 
out of domain generalization and wide applicability
\footnotetext[1]{https://tinyurl.com/4km4uwme} 
, and data augmentation is used to support analyses not covered in the original training data.
To qualify for practical 
use, we kept our methods simple, easy for deployment, and computationally modest 
(a single low-end GPU suffices for maximal performance). 
Thus, the demonstrated system represents a promising tool to leverage neural models for analysts routine work. 
To validate our fluency gains over the template-based methods, we showed that experts find n\textsc{BIIG} to produce insights that are more than three times as fluent as the template baseline, while still keeping faithfulness to the data very high.
The demo movie is available at \url{https://ibm.biz/BIIG_VIDEO}.

%% file: 3_system.tex
\begin{figure*}[t]
\begin{center}
  \includegraphics[width=0.85\textwidth]{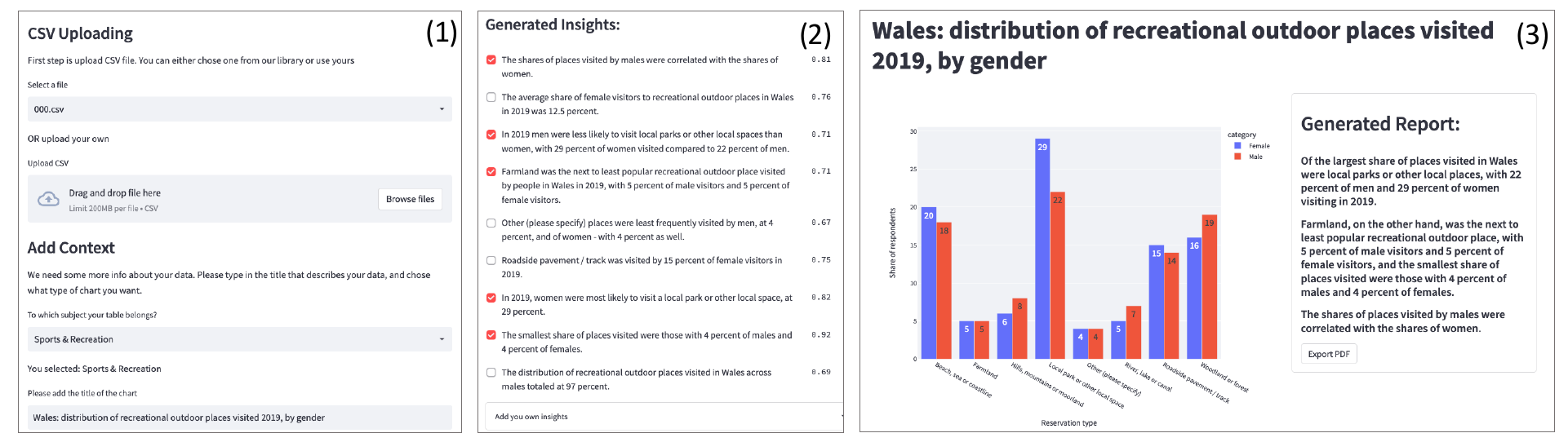}
 \end{center}
  \caption{Snapshots of the interaction- (1) Data and context; (2) Generated insights; (3) Table report.}~\label{fig:UI}
\end{figure*}

\begin{figure}[t]
\begin{center}
  \includegraphics[width=0.85\columnwidth]{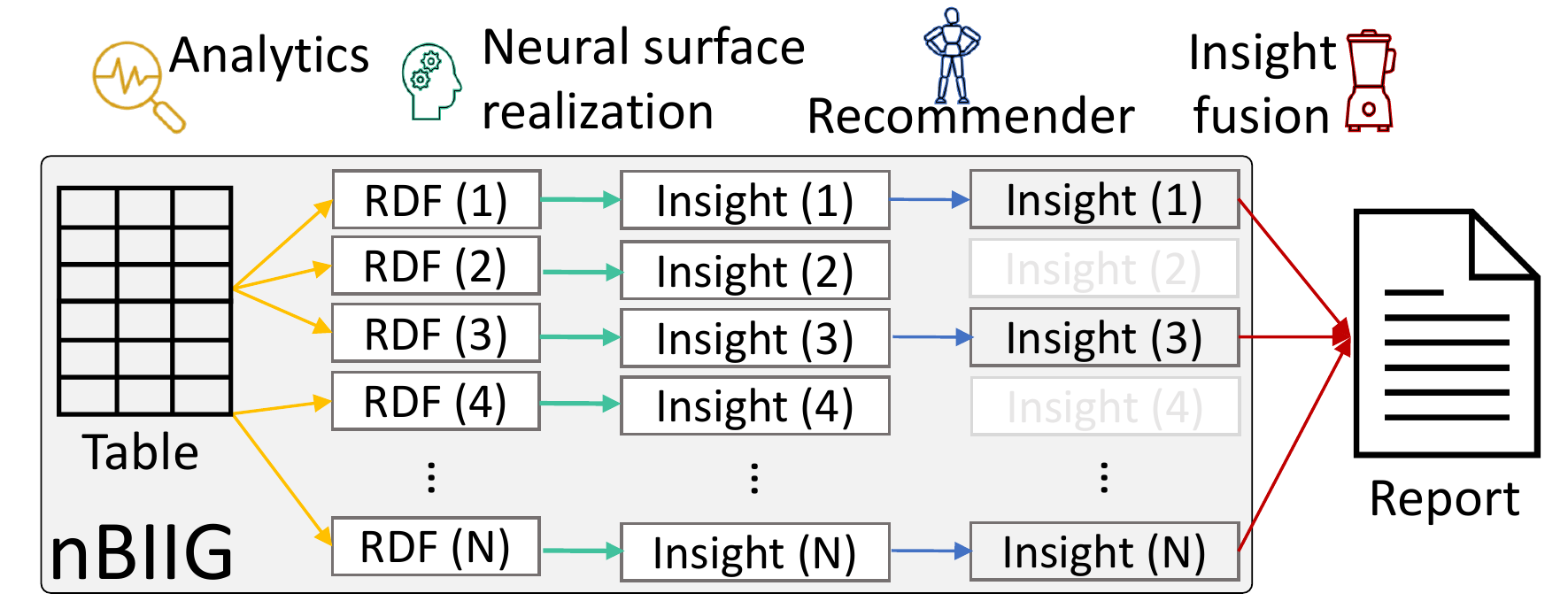}
 \end{center}
  \caption{n\textsc{BIIG} system flow and insight fusion module.}~\label{fig:system}
\end{figure}

\section{System Description and Quality Analysis}
\label{system}

n\textsc{BIIG}, illustrated in Figure~\ref{fig:system}, comprises three 
modules.\newline 
\textbf{Analytics:} The given table is fed into a suite of $10$ common 
BI analyses including MAX, MIN, SUM, AVERAGE, COMPARE, CORRELATED, RANKED, TREND, etc.
, using common statistical tools. Each relevant analysis result is cast into an RDF representation. 
\newline
\textbf{Neural Surface Realization:} We utilize T5~\cite{raffel2020exploring}, a transformer-based \textit{encoder-decoder} generative model, to translate the RDF representation for each analysis into a natural language insight. We fine-tune the model on data created using 
two different approaches. (1) An extractive approach applied to 100K publicly accessible webpages crawled from statista.com,
each 
containing a table 
and a summary written by a human expert. From every summary, we extracted each sentence that potentially conveys any of our supported analyses applied to the table, and paired the RDF cast for that analysis with the sentence.
We thus obtained 75K sentence/RDF matching pairs, with clear bias towards the simpler 
analysis types,
leaving very few examples for 
involved
types (e.g., RANKED, COMPARE). 
This scarcity
degrades the model performance on these low-prior types. 
(2) Data augmentation approach addresses that by utilizing GPT-NeoX-20B~\cite{gpt-neox-20b}, a Large Language Model (LLM). 
For each low-prior analysis type, 
we selected 2.5K RDFs of that type, that were not paired above,
and for each such RDF, generated a matching textual insight with the LLM using a few shot scheme. 
Specifically, we used a manually composed prompt 
of $5$ pairs of (RDF, textual insight) examples as context, 
followed by an RDF, 
for which we wish to obtain a textual insight.
The sequence generated by the model, given this input, was then taken to be that textual insight. 
\newline 
\textbf{Recommender:} 
To supply the analyst with insight recommendation, 
for each insight, we combine its faithfulness score obtained using~\cite{dusek-kasner-2020-evaluating}, a SOTA metric for semantic evaluation, together
with a score dependent on user preference logged in previous interactions.
This module improves flexibility, control, and explainability, all being 
key elements for BI solutions.
\newline 
To generate the report we offer an automatic \textbf{insights fusion} using rule-based heuristics which were shown useful in previous work, e.g., Project Debater~\cite{slonim2021autonomous}.
\\\newline
\textbf{Quality of Generated Insights   } 
To validate the quality of the generated insights, we conducted an evaluation by three NLP experts. We sampled $66$ tables and $529$ insights in total. Each insight was evaluated w.r.t faithfulness and fluency. 
For the $10$ supported types, on average $92\%$ were found to be faithful to the table, supporting the practical value of the system. For fluency, we additionally scored a commercial template-based insights generation system, we found that $93\%$ of insights produced by n\textsc{BIIG} were found to be fluent, a substantial gain over the $23\%$ of template-based methods.

%% file: 4_discussion.tex
\section{Demonstration and Discussion}

n\textsc{BIIG} user experience is shown in Figure~\ref{fig:UI}. Users interact with the system by uploading a table (CSV file), short relevant context, and the type of the chart required (line, column, etc.). Next, the user is presented with candidate 
insights suggested 
by the system, along with a score indicating the 
confidence in the 
faithfulness of the insights.
The user can then
edit or add new insights. Once ready, the analyst can select 
insights to be included in the report. Upon selection, a table report is generated, and the user can further edit the report or export it.
While the level of faithfulness attained by n\textsc{BIIG} outperforms previous works, a neural model is always prone to some level of errors. By indicating the faithfulness confidence score 
and facilitating editorial changes to the generated sentences, we enable
users to easily correct and customize the insights and the 
report, via an effective human-in-the-loop approach. 
In future work we intend to learn from users interactions to further improve the system quality. 

%% file: 7_acknowledgments.tex
\section{Acknowledgments}
We thank our colleagues at IBM Cognos Analytics: Obidul Islam, Don Banks, and Alain Chabrier, for valuable insights.

%% file: nbiig.bbl
\begin{thebibliography}{8}
\providecommand{\natexlab}[1]{#1}

\bibitem[{Black et~al.(2022)Black, Biderman, Hallahan, Anthony, Gao, Golding,
  He, Leahy, McDonell, Phang, Pieler, Prashanth, Purohit, Reynolds, Tow, Wang,
  and Weinbach}]{gpt-neox-20b}
Black, S.; Biderman, S.; Hallahan, E.; Anthony, Q.; Gao, L.; Golding, L.; He,
  H.; Leahy, C.; McDonell, K.; Phang, J.; Pieler, M.; Prashanth, U.~S.;
  Purohit, S.; Reynolds, L.; Tow, J.; Wang, B.; and Weinbach, S. 2022.
\newblock {GPT-NeoX-20B}: An Open-Source Autoregressive Language Model.
\newblock In \emph{Proceedings of the ACL Workshop on Challenges \&
  Perspectives in Creating Large Language Models}.

\bibitem[{Chen et~al.(2020)Chen, Chen, Zha, Zhou, Zhang, Sundaresan, and
  Wang}]{chen2020logic2text}
Chen, Z.; Chen, W.; Zha, H.; Zhou, X.; Zhang, Y.; Sundaresan, S.; and Wang,
  W.~Y. 2020.
\newblock Logic2text: High-fidelity natural language generation from logical
  forms.
\newblock \emph{arXiv preprint arXiv:2004.14579}.

\bibitem[{Du{\v{s}}ek and Kasner(2020)}]{dusek-kasner-2020-evaluating}
Du{\v{s}}ek, O.; and Kasner, Z. 2020.
\newblock Evaluating Semantic Accuracy of Data-to-Text Generation with Natural
  Language Inference.
\newblock In \emph{Proceedings of the 13th International Conference on Natural
  Language Generation}, 131--137. Dublin, Ireland: Association for
  Computational Linguistics.

\bibitem[{Perlitz et~al.(2022)Perlitz, Ein{-}Dor, Sheinwald, Slonim, and
  Shmueli{-}Scheuer}]{perlitzDiversity2022}
Perlitz, Y.; Ein{-}Dor, L.; Sheinwald, D.; Slonim, N.; and Shmueli{-}Scheuer,
  M. 2022.
\newblock Diversity Enhanced Table-to-Text Generation via Type Control.
\newblock \emph{CoRR}, abs/2205.10938.

\bibitem[{Raffel et~al.(2020)Raffel, Shazeer, Roberts, Lee, Narang, Matena,
  Zhou, Li, Liu et~al.}]{raffel2020exploring}
Raffel, C.; Shazeer, N.; Roberts, A.; Lee, K.; Narang, S.; Matena, M.; Zhou,
  Y.; Li, W.; Liu, P.~J.; et~al. 2020.
\newblock Exploring the limits of transfer learning with a unified text-to-text
  transformer.
\newblock \emph{J. Mach. Learn. Res.}, 21(140): 1--67.

\bibitem[{Ravichander et~al.(2019)Ravichander, Naik, Rose, and
  Hovy}]{ravichander2019equate}
Ravichander, A.; Naik, A.; Rose, C.; and Hovy, E. 2019.
\newblock EQUATE: A benchmark evaluation framework for quantitative reasoning
  in natural language inference.
\newblock \emph{arXiv preprint arXiv:1901.03735}.

\bibitem[{Slonim et~al.(2021)Slonim, Bilu, Alzate, Bar-Haim, Bogin, Bonin,
  Choshen, Cohen-Karlik, Dankin, Edelstein et~al.}]{slonim2021autonomous}
Slonim, N.; Bilu, Y.; Alzate, C.; Bar-Haim, R.; Bogin, B.; Bonin, F.; Choshen,
  L.; Cohen-Karlik, E.; Dankin, L.; Edelstein, L.; et~al. 2021.
\newblock An autonomous debating system.
\newblock \emph{Nature}, 591(7850): 379--384.

\bibitem[{Wang et~al.(2019)Wang, Sun, Zhang, Cui, Xu, Ma, and
  Zhang}]{wang2019datashot}
Wang, Y.; Sun, Z.; Zhang, H.; Cui, W.; Xu, K.; Ma, X.; and Zhang, D. 2019.
\newblock Datashot: Automatic generation of fact sheets from tabular data.
\newblock \emph{IEEE transactions on visualization and computer graphics},
  26(1): 895--905.

\end{thebibliography}
